\documentclass[runningheads]{llncs}

\usepackage{eccv}

\usepackage{eccvabbrv}
\usepackage{graphicx}
\usepackage{booktabs}
\usepackage{multirow}
\usepackage{wrapfig}
\usepackage{float}

\usepackage[accsupp]{axessibility}  

\usepackage{hyperref}

\usepackage{orcidlink}

\begin{document}

\title{Human Pose Estimation in Trampoline Gymnastics: How to Improve Performance on Extreme Poses}

\titlerunning{HPE in Trampoline: How to Improve Performance}

\author{Léa Drolet-Roy\thanks{Equal contribution.}\inst{1,2}\orcidlink{0009-0004-1985-5006} \and
Victor Nogues$^{\star}$\inst{1}\orcidlink{0009-0004-6234-0729} \and
Bérenger Chedal-Anglay\inst{1} \and
Sylvain Gaudet\inst{2} \and
Eve Charbonneau\inst{3}\orcidlink{0000-0002-9215-3885} \and
Mickaël Begon\inst{4}\orcidlink{0000-0002-4107-9160} \and
Lama Séoud\inst{1}}

\authorrunning{L.~Drolet-Roy et al.}

\institute{Polytechnique Montreal, Canada\\
\email{\{lea.drolet-roy,victor.nogues\}@etud.polymtl.ca}\\
\and Institut National du Sport du Québec, Canada \and
University of Sherbrooke, Canada \and
University of Montreal,  Canada}
\maketitle

\begin{abstract}
  Trampoline gymnastics involves extreme human poses and uncommon viewpoints, on which state-of-the art pose estimation models tend to under-perform. We demonstrate that this problem can be addressed by fine-tuning a pose estimation model on a combination of real extreme poses and domain-specific synthetic poses (STP). We generate STP from motion capture recordings of trampoline routines. We propose a pipeline to fit noisy motion capture data to a parametric human model, then generate multi-view realistic images with high-fidelity keypoint labels. The fine-tuned ViTPose model tested on real multi-view images exhibits accuracy improvements in 2D which translate to improved 3D triangulation. In 2D, we obtain a performance similar to state-of-the-art models on the MS COCO validation set while evaluating on significantly more challenging data, bridging the performance gap between common and extreme poses. In 3D, we reduce the MPJPE by 46.1~mm with our best model, which represents an improvement of 42.7\% compared to the pretrained ViTPose model. Our code and data are available at \href{https://github.com/VisionICLab/trampoline_syn_data}{https://github.com/VisionICLab/trampoline\_syn\_data}.
  
\end{abstract}

\begin{figure}[h]
    \centering
    \includegraphics[width=\textwidth]{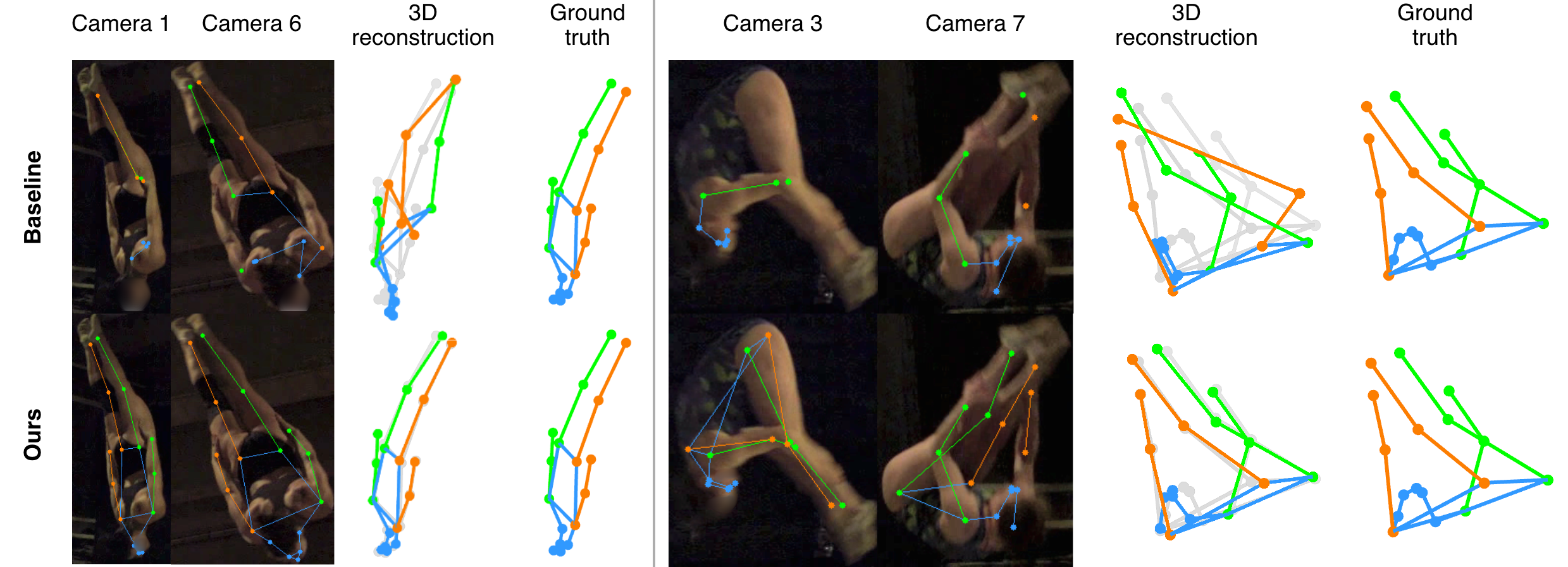}
    \caption{Qualitative comparison of the ViTPose-S without \textit{(Baseline)} and with fine-tuning \textit{(Ours)} in extreme cases, 2 views out of 8 are selected. Illustrated triangulations are obtained from 8 views, using Pose2Sim\cite{pagnon_pose2sim_2022} weighted triangulation function. Ground-truth poses are shown in gray behind the reconstructed poses.}
    \label{fig:intro}
\end{figure}

\section{Introduction}
\label{sec:intro}
Understanding human motion is fundamental to applications ranging from sport analytics to animation. Among these scenarios, acrobatic motions such as those performed in trampoline gymnastics represent some of the most challenging forms of human movement. During trampoline routines, athletes execute combinations of somersaults and twists, performed in tucked, piked and straight positions. This involves atypical poses with extreme joint angles, strong self-occlusions and rapid body orientation changes leading to motion blur. These challenges are even tangible in manual pose annotation, emphasizing the complexity of the task. Accurate pose estimation in such scenarios can enable quantitative analysis of sport performance, new forms of markerless motion capture for sports and automatic scoring, like demonstrated in a pilot project by Fujitsu in artistic gymnastics \cite{fujiwara2018ict}. However, despite recent progress in human pose estimation (HPE), current solutions remain unreliable for such extreme motions, as depicted in \cref{fig:intro}, largely because they are trained on datasets dominated by upright everyday activities \cite{lin_microsoft_2015, ionescu_human36m_2014, mahmoodAMASSArchiveMotion2019}. Creating datasets that capture these challenging cases is difficult. Marker-based motion capture systems can record complex movements with high-precision, but trampoline sequences suffer from marker detachment and occlusions, making it hard to directly reconstruct consistent body motion.

In this work, we address these challenges by introducing a pipeline for improving HPE in trampoline gymnastics. The pipeline begins with marker-based motion capture recordings of trampoline routines. Because these sequences frequently suffer from markers detachment or occlusions, we propose a method for fitting a parametric human body model (from the SMPL \cite{loperSMPLSkinnedMultiperson2015} family) to noisy motion capture sequences. We then leverage the recovered motions to generate a synthetic multi-view dataset with diverse clothing, backgrounds and camera viewpoints, with accurate 2D and 3D pose annotations. This synthetic dataset is used to fine-tune a ViTPose\cite{xu_vitpose_2022} model. Finally, we evaluate 2D pose estimation and multi-view 3D pose reconstruction via triangulation on real multi-view trampoline routines recordings.

Our contributions are:
\begin{enumerate}
    \item A new method to fit a SMPL body-model to noisy and complex motion capture data.
    \item A collection of SMPL trampoline motion sequences featuring real-world extreme human poses under-represented in existing datasets such as AMASS.
    \item A new end-to-end open-source tool for HPE synthetic data generation.
    \item A ViTPose model fine-tuned on trampoline motion sequences and real sports data \cite{johnson_learning_2011} that achieves SOTA HPE performance on real-world complex acrobatic data, both on 2D and triangulated 3D joints positions.
\end{enumerate}

More broadly, the challenges addressed in this paper, namely markerless motion capture under fast and atypical movement, robustness to self-occlusion, and the domain gap between everyday motion training data and rare movement patterns, are shared by other real-world human motion settings. Even for clinical gait and movement analysis, standard pose estimators are similarly trained on datasets dominated by typical and unoccluded motion.

\section{Related works}
\subsection{Human pose estimation}

Most HPE pipelines decompose the problem into detecting 2D keypoints in images, optionally followed by 3D reconstruction using either monocular inference or multi-view geometry. A first category of HPE approaches relies on CNN for detecting 2D keypoints like OpenPose \cite{cao2019openpose} and RTMpose \cite{jiang2023rtmpose}. The second one includes transformer-based architectures. In particular, ViTPose \cite{xu_vitpose_2022} leverages Vision Transformer backbones to achieve state-of-the-art performance on large-scale benchmarks such as COCO, demonstrating the effectiveness of large scale pretraining combined with fine-tuning strategies. 

Recovering 3D human pose from multiple synchronized cameras provides a reliable alternative to monocular 3D HPE by leveraging geometric consistency across views. Independently detected 2D keypoints in each camera view are used to reconstruct 3D keypoints through geometric triangulation. Several works have demonstrated that accurate multi-view reconstruction can be achieved when high-quality 2D detections are available and cameras are properly calibrated \cite{pavlakos2017harvesting, dong2019fast, bartol2022generalizable}. However, the performance of these systems strongly depends on the robustness of underlying 2D pose estimators. In challenging scenarios such as acrobatic movements, self-occlusions and uncommon viewpoints can significantly degrade the quality of 2D keypoint predictions, thereby reducing the accuracy of the reconstructed 3D pose. These situations are rarely represented in existing datasets, motivating the development of specialized datasets and synthetic data generation strategies to improve robustness in such settings.

\subsection{Human motion capture datasets}

Large-scale human motion capture datasets have played a pivotal role in advancing research in HPE and motion analysis. A widely used resource is the AMASS collection \cite{mahmoodAMASSArchiveMotion2019}, which aggregates motion capture recordings from multiple datasets into a unified representation using the SMPL body model. While AMASS contains a broad range of activities, the large majority of recorded sequences correspond to everyday movements. Among the datasets included in AMASS, MoYo \cite{tripathi_intuitive_physics_2023} introduces motions with larger joint angles ranges and more dynamic movements. However, even this dataset remains limited in its coverage of highly dynamic acrobatic movements involving frequent body inversion and rapid rotations. Such motions are difficult to capture reliably with traditional marker-based motion capture systems due to occlusions, marker detachment and the high-speed nature of the motion. Camera positioning is also a challenge, as we need to cover jumps up to eight meter high.

In this work, we focus specifically on acrobatic trampoline motion, a class of movements largely absent from existing motion capture collections. To address the challenges associated with capturing such motion, we propose a method for fitting a SMPL model to noisy motion capture data.

\subsection{SMPL body-model fitting}

Transferring human motion to animatable avatars is a widely explored field as several works have focused on recovering SMPL parameters from marker-based motion capture data. Early approaches such as MoSh \cite{loperMoShMotionShape2014} and MoSh++ \cite{mahmoodAMASSArchiveMotion2019} enabled large-scale unification of heterogeneous motion capture datasets resulting in the AMASS collection. More recent works, like SOMA \cite{ghorbani_soma_2021} or DAMO \cite{kimDAMODeepSolver2025}, have explored learning-based approaches to improve robustness to noisy or incomplete marker data. While these approaches significantly improve automation of motion capture processing, they are typically designed for conventional scenarios and standard human motions. In contrast, acrobatic movements introduce additional challenges, including rapid motion and frequent marker detachment. Methods like MoSh \cite{loperMoShMotionShape2014} or SOMA \cite{ghorbani_soma_2021} do not handle these cases well, produce unrealistic, distorted outputs, and are extremely slow to run on large batches of data. In this work, we address these challenges by proposing a robust GPU-accelerated SMPL fitting method, tailored to noisy motion capture data of trampoline routines.

\subsection{Synthetic human pose datasets}

Synthetic data generation has become common practice in HPE, with large datasets like SURREAL \cite{varolLearningSyntheticHumans2017}, AGORA \cite{patelAGORAAvatarsGeography2021a} and BEDLAM \cite{blackBEDLAMSyntheticDataset2023, tesch2025bedlam2}. These works demonstrate that combining synthetic and real data improves HPE as synthetic datasets enable greater diversity through variations in body shape, appearance, motion and environments, as well as the inclusion of atypical poses and uncommon camera configurations that are difficult to capture or annotate in real-world. 

However, specific contexts remain under-represented in existing datasets. For example, scenarios involving people in wheelchairs are not adequately covered in standard HPE datasets. To address such gaps, recent works such as WheelPose \cite{huang_wheelpose_2024} and RePoGen \cite{purkrabek_improving_2024} propose specialized synthetic datasets designed to improve HPE in specific settings. In contrast, our work focuses on generating synthetic training data derived from real-world trampoline motion capture, enabling HPE under extreme acrobatic poses.

\section{Methods}
\subsection{SMPL body-model fitting to motion-capture data}

With athletes moving fast and experiencing large acceleration forces, the markers used for traditional motion capture tend to detach from the athletes, leading to incomplete marker data, or diverging trajectories. 
Our primary goal is not to match exactly a set of markers, but to target smooth, plausible, complex motion that accounts for missing or diverging markers. 
To retrieve realistic motion from a set of markers, we make the following prior assumptions:
\begin{itemize}
    \item markers order remains constant over time, indicating that the acquisition system maintains consistent marker identities throughout the sequence;
    \item a user-defined marker layout specifying the association between markers and body segments is available a priori;
    \item markers attached to the same rigid body segment should maintain fixed pairwise distances over time.
\end{itemize}

Assuming these hypotheses are valid, and before starting the fitting, we automatically filter the raw marker sequences to remove unreliable trajectories. The procedure discards markers with invalid labels or high residual errors (from the motion capture system), non-moving markers, and those whose vertical dynamics deviate from the dominant jump pattern observed across markers. This last part is specific to trampoline data, and can be deactivated. It is also important to note that the user-defined marker layout is only required once per marker set, allowing us to set these correspondences once to fit ten sequences executed by three different athletes. As motion capture data collections usually employ well-defined and unchanging marker sets, our method is suitable for large-scale dataset generation.

\begin{figure}[H]
    \centering
    \includegraphics[width=0.97\textwidth]{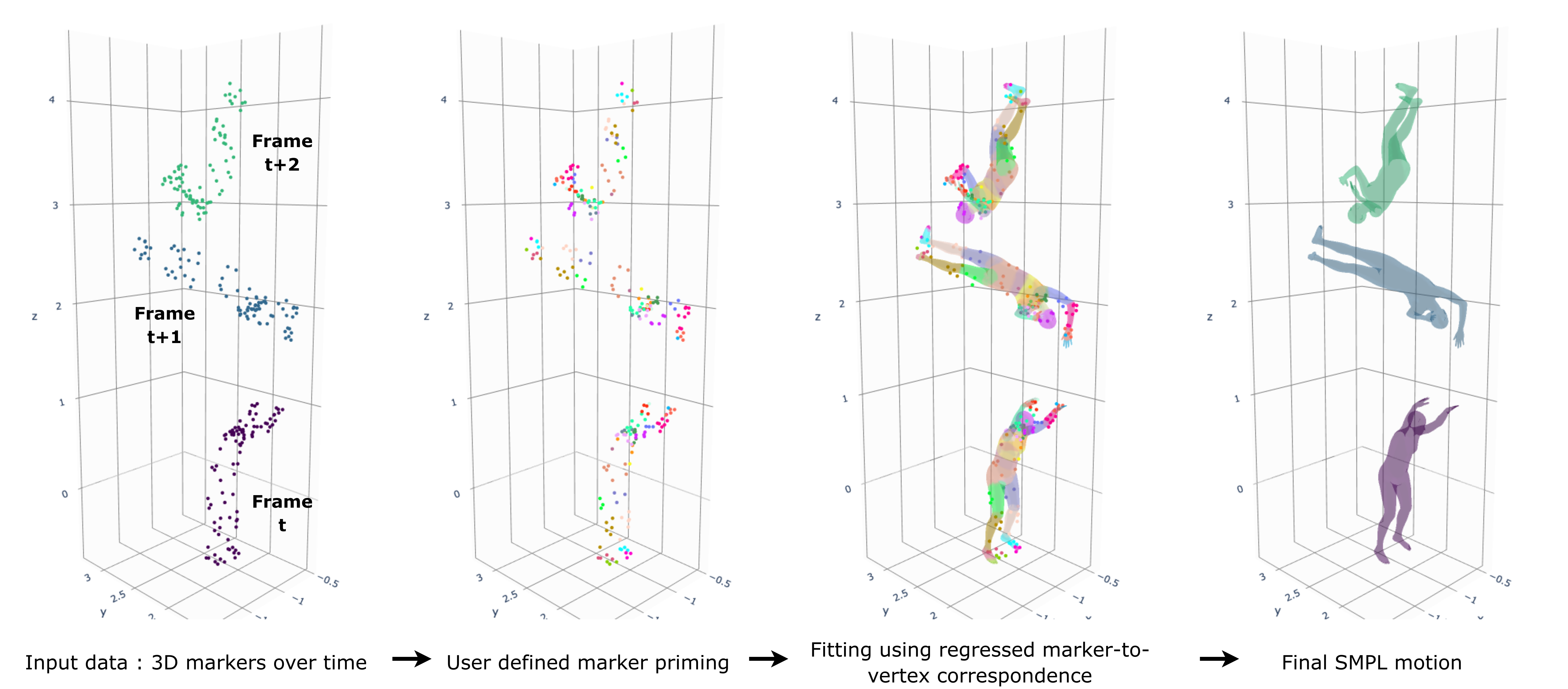}
    \caption{Markers-to-SMPL fitting pipeline. Illustrated frames are not consecutive.}
    \label{fig:bodyparts}
\end{figure}

We then proceed to a rough marker-to-vertex fitting process on a few poses sampled uniformly in the target sequence. The synthetic markers, positioned 9.5 mm above the skin surface, as the real markers were, are expected to match the observed marker locations. We estimate the marker–vertex correspondence by accumulating, over a downsampled sequence, the number of frames in which each marker projects onto each vertex, and assigning the marker to the vertex with the highest support. A visual evaluation tool is provided to qualitatively determine if the markers placement is plausible. \cref{fig:bodyparts} illustrates the fitting process, from filtered marker coordinates to final SMPL fittings.

Once a satisfactory vertex-to-marker correspondence set is found, a marker-to-synthetic-marker distance loss is applied with temporal smoothing and Gaussian mixture model regularization, used also in SMPLify \cite{bogoKeepItSMPL2016}, to avoid converging toward unrealistic poses. The fitting phase is fully customizable, although we used a succession of four steps during our experiments:
\begin{itemize}
    \item LBFGS \cite{liu_limited_1989} optimization only focused on translation and orientation,
    \item Adam optimization only focused on translation and orientation refinement,
    \item short Adam optimization phase to regress pose, shape, translation and orientation.
    \item LBFGS optimization to refine pose, shape, translation and orientation.
\end{itemize}

The output is a smooth, realistic SMPL sequence that closely follows the original motion, while ignoring detached markers. We compare our approach to SOMA on four sequences from three athletes. SOMA produced unrealistic results, including physically implausible joint angles, whereas our method yields smooth realistic motion and reduces computation time by 57–70\%, depending on the marker layout and sequence length. \cref{tab:soma} provides quantitative metrics.

\begin{table}[h]
\caption{Marker-to-closest-vertex after fitting. SOMA uses SMPLX, whereas our method uses SMPL. Standard deviation is provided for the error. Both methods were run on the same machine equipped with an Intel Core i9-13900 CPU and an NVIDIA GeForce RTX 4090 GPU.}
\centering
\setlength{\tabcolsep}{8pt}
\begin{tabular}{cc|cc|cc}
\toprule
 & & \multicolumn{2} {c|} {Error (mm) $\downarrow$} & \multicolumn{2} {c} {Processing time $\downarrow$}\\
Subject ID & \# frames & SOMA  & Ours & SOMA & Ours \\ \midrule
1 &3347&    $41.1 \pm 54.5$ &\textbf{28.1} $\pm$ \textbf{30.2} & 42m 27s & \textbf{15m 10s}\\
2 &2412&    $36.4 \pm 25.9$ &\textbf{17.3} $\pm$ \textbf{12.0} &  37m 01s & \textbf{10m 56s}\\
3a &5079&   $41.3 \pm 44.7$ &\textbf{22.7} $\pm$ \textbf{23.5} & 49m 54s & \textbf{21m 27s} \\
3b &4537&   $30.9 \pm 25.3$ &\textbf{16.2} $\pm$ \textbf{9.3} & 45m 23s& \textbf{19m 07s} \\

\bottomrule
\end{tabular}
\label{tab:soma}
\end{table}

Our SMPL fitting pipeline is not specific to sport and could equally be applied to generate domain-specific fine-tuning data for other under-represented, real-world motion distributions.

\subsection{TramPoseFit motion dataset}
Our SMPL fitting method is applied to motion capture data of ten trampoline sequences executed by three elite athletes \cite{venne_optimal_2023}. Each sequence contains one to five acrobatic jumps. For this data collection, a total of 95 reflective markers were placed on trampolinists. Markers positions were captured using 18 Vicon optoelectronic cameras at 200 FPS. The filtering process ended up only using a mean number of $59\pm19$ markers (about 62\% of the whole 95 markers set). We refer to the resulting motion collection as TramPoseFit.

\subsection{SynTramPose multi-view synthetic dataset (STP)}
With TramPoseFit, we proceed to generating realistic multi-view synthetic images using Blender (Stitching Blender Foundation, Amsterdam).
First, a camera combination is chosen: here, we define an eight-camera setup similar to the one used during our real-world acquisitions, illustrated on \cref{fig:cameras}, and another setup with modified camera viewpoints. The cameras were arranged to ensure a wide range of viewpoints while maintaining sufficient coverage of the trampoline volume. The TramPoseFit avatar performing the motion is then placed in the center of the world reference frame.

An HDRI background is then added to the scene to ensure realistic surroundings around the avatar. Three different HDRI backgrounds (under CC0 license) are used. Their combination represents common environment challenges such as low contrast and luminosity, multiple background objects, or vibrant colors. A body texture from BEDLAM \cite{blackBEDLAMSyntheticDataset2023} is applied to the avatar. Avatars skin textures are randomly sampled and modified for each camera view. A simple clothing set is added by selecting a submesh of the SMPL avatar, and expanding this mesh to place its surface a few millimeters away from the body. Colors are then applied to the clothing submesh to distinguish it from the skin. We do not add any physics-based deformation due to the tightness of actual athletes clothing.

\begin{figure}[h]
    \centering
    \begin{minipage}[t]{0.9\textwidth}
        \centering
        \begin{minipage}[l]{0.24\textwidth}
            \includegraphics[width=\textwidth]{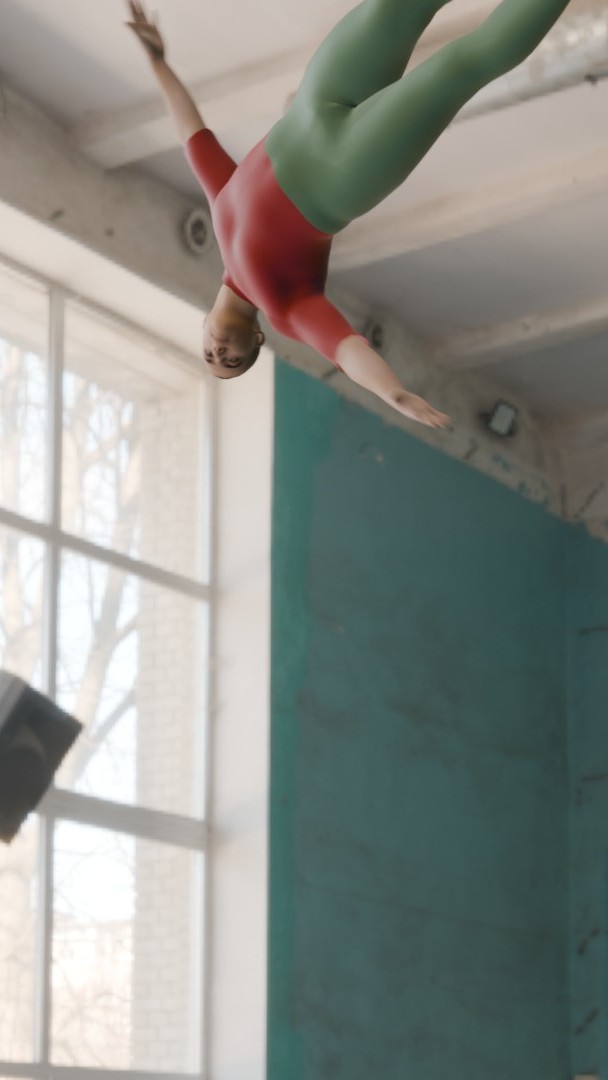}
        \end{minipage}
        \begin{minipage}[c]{0.24\textwidth}
            \includegraphics[width=\textwidth]{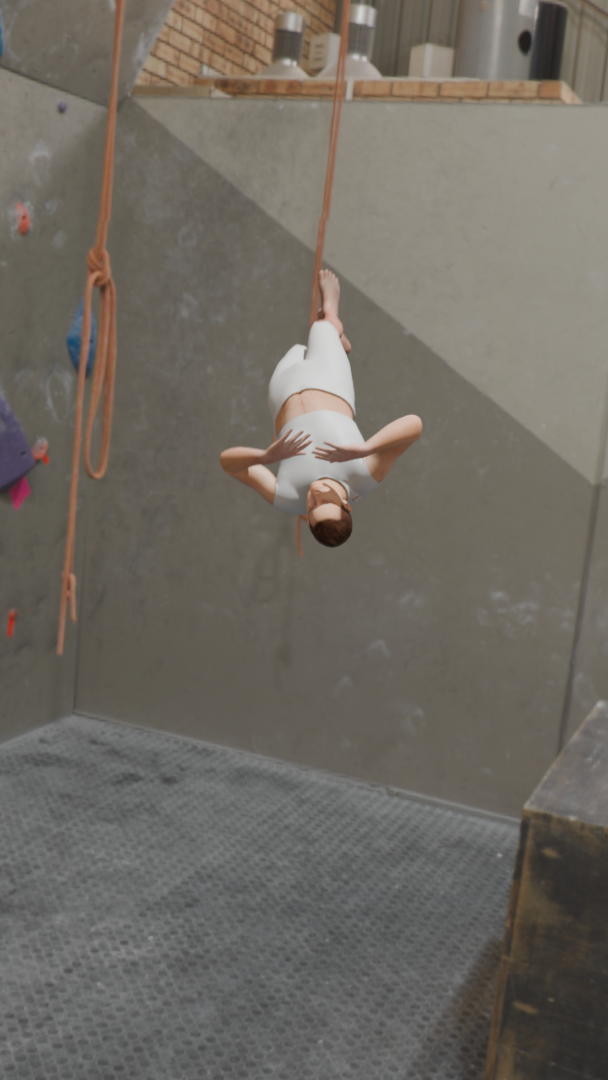}
        \end{minipage}
        \begin{minipage}[c]{0.24\textwidth}
            \includegraphics[width=\textwidth]{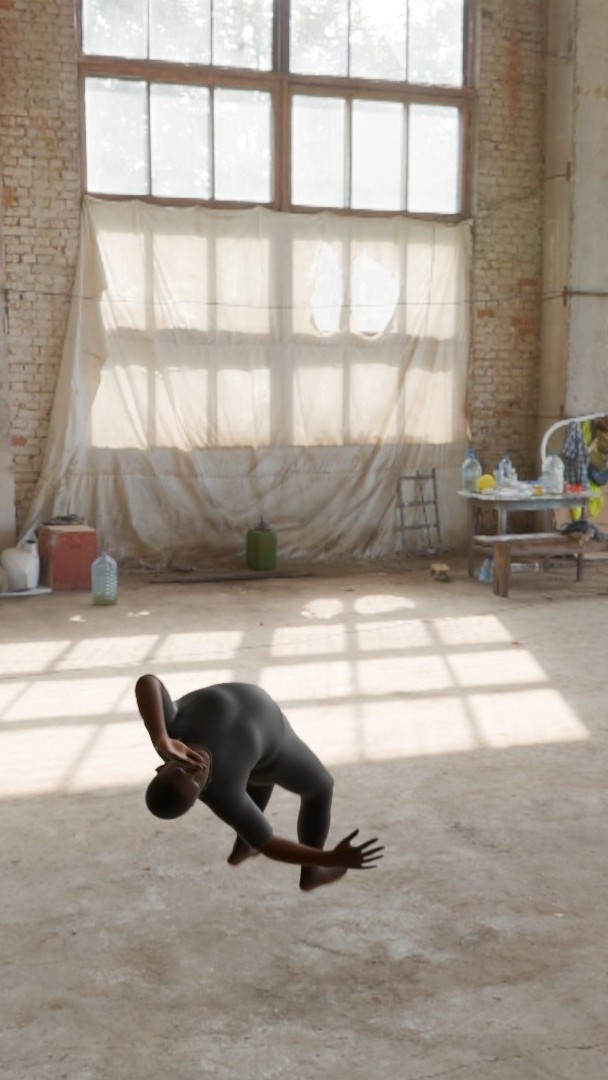}
        \end{minipage}
        \begin{minipage}[r]{0.24\textwidth}
            \includegraphics[width=\textwidth]{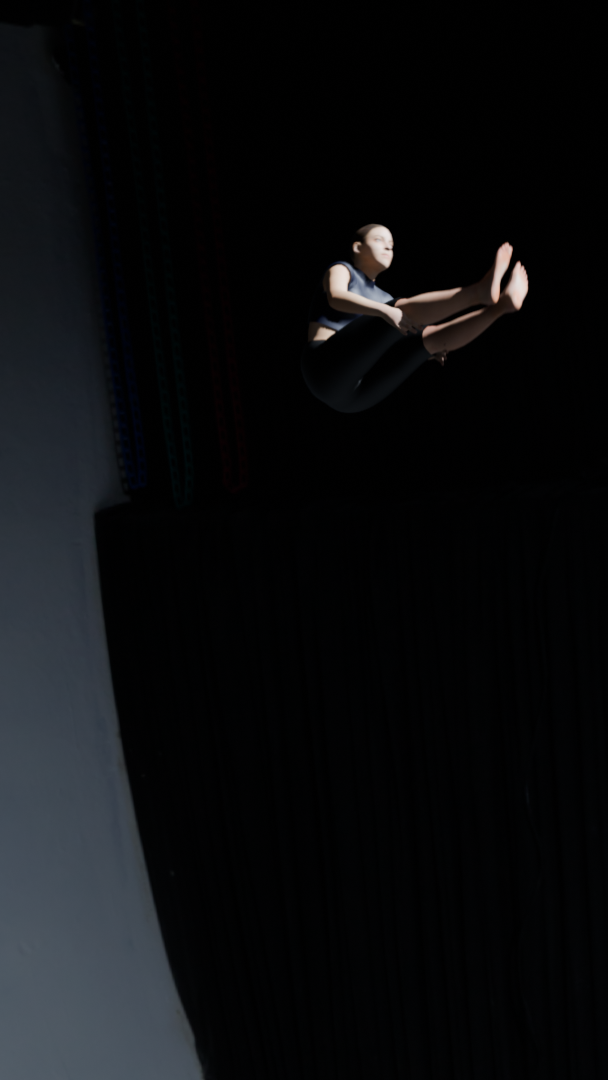}
        \end{minipage}
    \end{minipage}
    \caption{Examples of synthetic images in STP (images are cropped for clarity)}
    \label{fig:STPexamples}
\end{figure}

The resulting SynTramPose (STP) dataset contains a total of 3,624 synthetic images and their corresponding 2D joint automatic annotations. Sample images are provided in \cref{fig:STPexamples}. We cover a diversity of camera views, backgrounds, avatar skin textures and clothes, as explained in the pipeline. To avoid any form of data leakage athletes visible in SRT (validation set) or in MRT (test set) and also present in TramPoseFit are excluded from STP (ID 3). STP is then generated from six sequences executed by two athletes.

\textbf{Joint annotations.} We generate automatic joint annotations in the format [x, y, v] where v stands for visibility. Following COCO format guidelines \cite{lin_microsoft_2015}, v has three possible values: v=0 if not visible, v=1 if labeled but occluded and v=2 if labeled and visible. We define the joint visibility first by verifying if the joint position falls within the image bounds. If not, v=0. Subsequently, we verify if the joint is occluded or visible, using the method detailed by \cite{purkrabek_improving_2024}.

\section{Experimental setup}

\subsection{Implementation details}

We considered the ViTPose-Small model \cite{xu_vitpose_2022} to allow comparison with prior work \cite{purkrabek_improving_2024} while considering the trade-off between computational cost and performance. We use a batch size of 256 and a base learning rate of 1e-4 with the AdamW optimizer \cite{loshchilovDecoupledWeightDecay2019}. We use the OpenMMLab MMPose toolbox for conducting our experiments \cite{mmpose_contributors_openmmlab_2020}. We consider the models pre-trained on the COCO dataset as our baseline models. Then, we fine-tune a model on extreme poses, including our STP dataset and a publicly available dataset of real images of extreme sports poses (LSPe\cite{johnson_learning_2011}). 

\subsection{Datasets}

Experiments are conducted using the following datasets. In all datasets, we crop the images to ground truth bounding boxes, since our goal is to evaluate only the pose estimator, excluding detection errors. Manual annotation is too time-consuming to consider annotating a dataset for model training, although it is reasonable to thoroughly annotate subsets for model validation and testing. The manual annotation method is described at the end of this section.

\textbf{Leeds Sports Pose extended (LSPe)} \cite{johnson_learning_2011}. LSPe contains 10,000 real-world sports images including gymnastics and parkour, where poses resemble trampoline ones. 2D HPE annotations are gathered through Amazon Mechanical Turk. LSPe is considered in our experiments to mitigate the domain gap between synthetic and real images, without harming performance on extreme poses. It is optionally used for fine-tuning.

\begin{figure}[H]
    \centering
    \begin{minipage}[r]{0.82\textwidth}
        \begin{minipage}[l]{\textwidth}
            \includegraphics[width=0.12\textwidth]{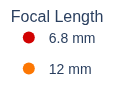}
        \end{minipage}
        \begin{minipage}[b]{\textwidth}
            \centering
            \includegraphics[width=\textwidth]{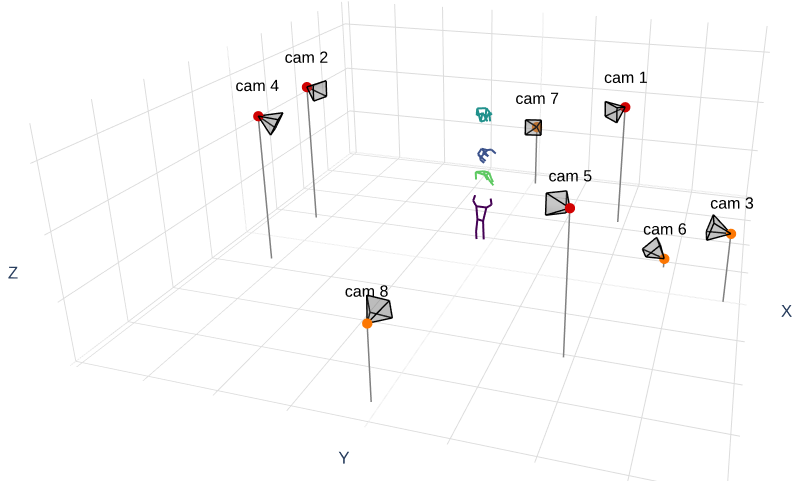}
        \end{minipage}
    \end{minipage}
    \caption{Multi-view acquisition setup. A trampolinist in motion is represented as a skeleton at different positions.}
\label{fig:cameras}
\end{figure}

\textbf{Singleview Real TramPose (SRT)}. SRT is a private manually annotated dataset of real trampoline recordings. It consists of a combination of two subsets: a) eight jumps executed by an elite trampolinist and captured with a webcam at 30 FPS (103 images) and b) one additional jump executed by the same trampolinist and captured with eight OptiTrack Prime Color cameras at 120 FPS and downsampled at 60 FPS for annotation (746 images). The OptiTrack cameras were synchronized and calibrated. The eight-camera setup is shown on \cref{fig:cameras}. The camera positions were chosen so that every point of the acquisition volume was seen by at least three cameras to facilitate the triangulation process. Constraints on the security distance around the trampoline were taken into account, which explains the limited size of athletes in the images. Cameras were calibrated using a traditional calibration method \cite{zhang_flexible_2000} and a ChArUco board, using OpenCV library\cite{noauthor_opencv_nodate}. After calibration, triangulating a point from an image to the world reference frame and projecting it back on the image gives a RMSE of 1.1 pixels. Data collection was approved by the Research Ethics Committee of our university and all participants signed written consents. This dataset is used only for model validation.

\textbf{Multiview Real TramPose (MRT)}. MRT is a private manually annotated dataset of real multi-view trampoline recordings. The RGB images were acquired using eight OptiTrack Prime Color cameras running at 120 FPS, during trampolinists training sessions, using the setup presented on \cref{fig:cameras}. MRT is a combination of two subsets: MRTa corresponds to the recordings of two jumps executed by two different athletes (292 images) and MRTb corresponds to two consecutive jumps executed by the same athlete (264 images). All sequences are downsampled to 12 FPS to ensure enough pose variation between images. MRT is only used for model testing. It allows a 2D error quantification in different views as well as an extensive 3D error quantification through multi-view triangulation.

\subsection{Manual annotations}

Two different GUIs are used for annotating: Label Studio and MV-PoseAnnotator\footnote[1]{Available at \href{https://github.com/VisionICLab/MV-PoseAnnotator}{https://github.com/VisionICLab/MV-PoseAnnotator}} 

(MVPA), a home-made tool that allows to assess multi-view consistency through triangulation and reprojection. SRT and the first part of MRT are annotated with Label Studio, while the second part of MRT is annotated using MVPA. When using MVPA, the keypoints were positioned on three selected views, triangulated, reprojected on all views, and then rectified to ensure both image-level accuracy and multi-view consistency. MVPA is helpful in ambiguous cases because the other views help the annotator understand where a given joint is. The triangulation-and-reprojection functionality makes the annotation process faster and more consistent between views.

\begin{table}[h]
\caption{Summary of manually annotated datasets details}
\centering
\setlength{\tabcolsep}{8pt}
\begin{tabular}{lccccc}
\toprule
Dataset  & Athlete(s) ID & \# images & FPS & Annotation tool \\
\midrule
SRTa     & 3             & 103      & 30        & Label Studio           \\
SRTb     & 3             & 746      & 60        & Label Studio           \\
MRTa     & 4, 5          & 292      & 12        & Label Studio           \\
MRTb     & 5             & 264      & 12        & MVPA                   \\
\bottomrule
\end{tabular}
\label{tab:datasets}
\end{table}

\subsection{2D HPE performance against baseline}

We first compare the fine-tuned models with the baseline models in terms of 2D HPE performance on MRT test set, following COCO guidelines \cite{lin_microsoft_2015}. Reported average precision (AP) and average recall (AR) are OKS-based. Results are provided in \cref{tab:AP}. Baseline corresponds to COCO pre-trained ViTPose and RTMPose models, with available checkpoints. ViTPose is chosen for fine-tuning as it is the best baseline. The fine-tuning is conducted using 5,000 common images from COCO2017-val in addition to real images of uncommon sports poses (LSPe) images and/or synthetic trampoline images (STP). Individually, both new datasets demonstrate an accuracy gain compared to the baseline model, but the best performance is reached when combining both datasets. Our model achieves comparable performance to ViTPose-S on the MS COCO validation set (AP=73.8, AR=79.2 as reported in \cite{xu_vitpose_2024}) while being evaluated on significantly more challenging data. Assuming ViTPose can be considered as an established solution for standard 2D HPE, these results suggest that the fine-tuned model provides a practical solution for pose estimation in trampoline scenarios.

\begin{table}[h]
    \caption{Performance of pose estimators pre-trained on COCO and fine-tuned on COCO (5,000 images) and additional datasets - testing on MRT. The results are the mean performance with 5 different seeds. (*) means that we did not train on RePoGen, but only use the provided weights\cite{purkrabek_improving_2024}}
    \label{tab:AP}
    \centering
    \setlength{\tabcolsep}{7pt}
    \begin{tabular}{@{}clccccc@{}}
        \toprule
        Model     & Fine-tuning data & N images & AP   & AP 50 & AR   & AR 50 \\
        \midrule
        \multirow{1}{*}{RTMPose-S} & None (\textit{Baseline})      &     -  & 40.1 & 69.8 & 47.9 & 77.4 \\
        \midrule
        \multirow{5}{*}{ViTPose-S} & None (\textit{Baseline})      &     -  & 50.5 & 72.9 & 55.6 & 77.9 \\
                                & STP                              &  3,624    & 59.3 & 84.6 & 64.0 & 88.0 \\
                                & LSPe                              &  10,000  & 64.1 & 87.4 & 70.2 & 90.8 \\
                                & LSPe + STP \textit{(Ours)}      &  13,624    & \textbf{68.4} & \textbf{88.0} & \textbf{74.5} & \textbf{91.7} \\
                                & RePoGen *          & -     & 59.1 & 80.9 & 70.0 & 88.7 \\
        \bottomrule
    \end{tabular}
\end{table}

\begin{figure}[h!]
    \begin{minipage}[l]{0.25\textwidth}
        \includegraphics[width=\textwidth]{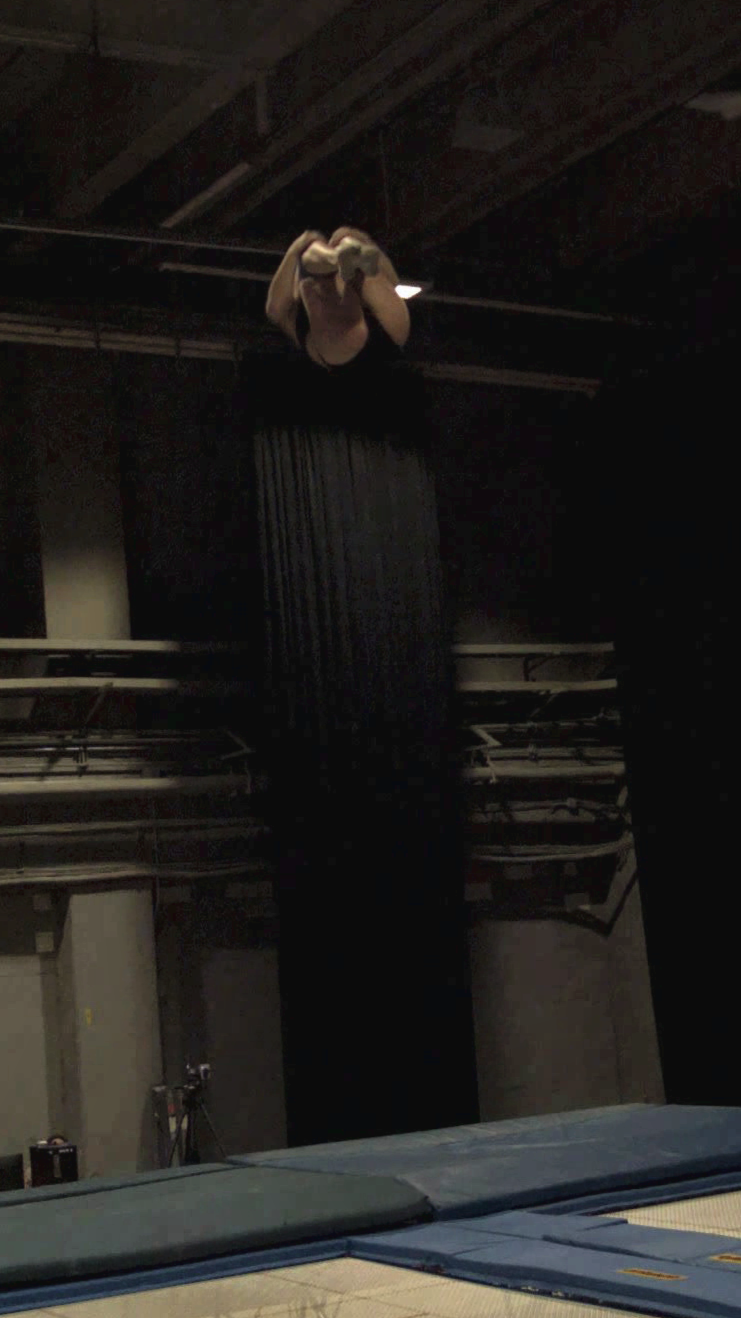}
    \end{minipage}
    \centering
    \begin{minipage}[c]{0.25\textwidth}
        \includegraphics[width=\textwidth]{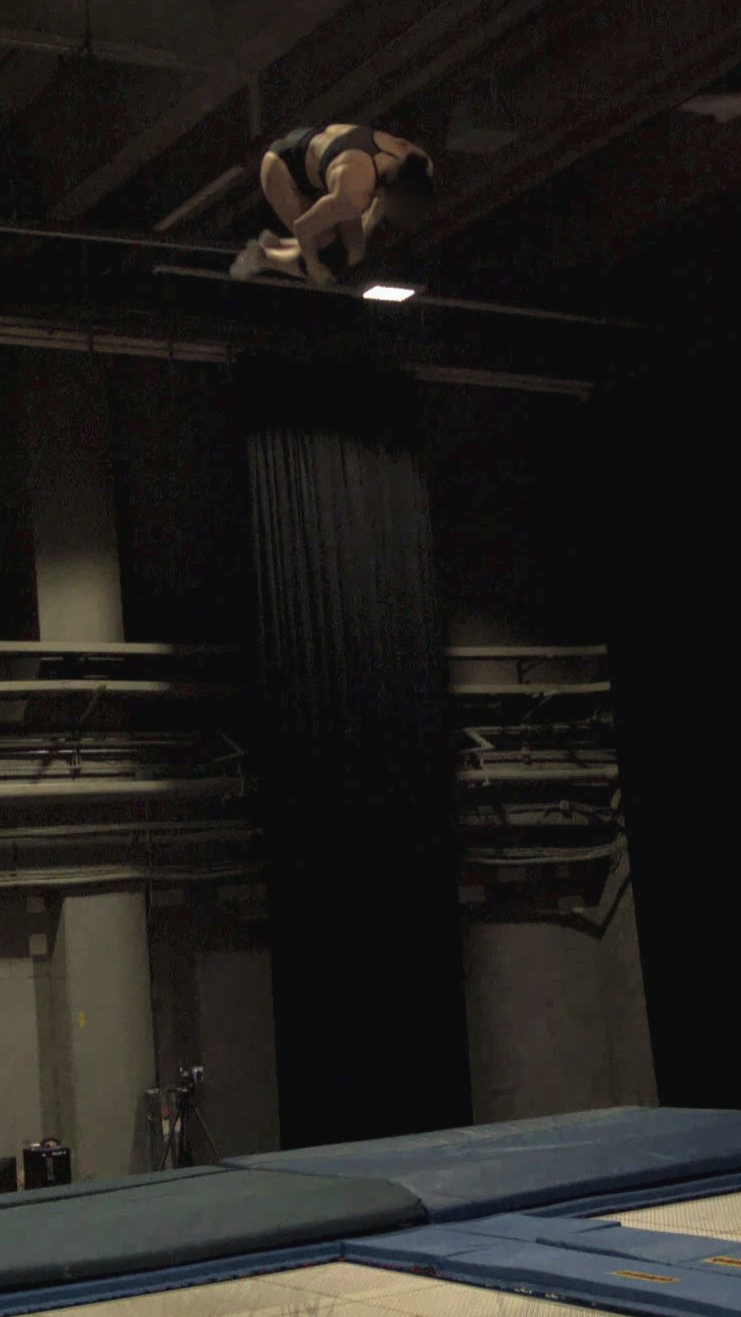}
    \end{minipage}
    \begin{minipage}[r]{0.25\textwidth}
        \includegraphics[width=\textwidth]{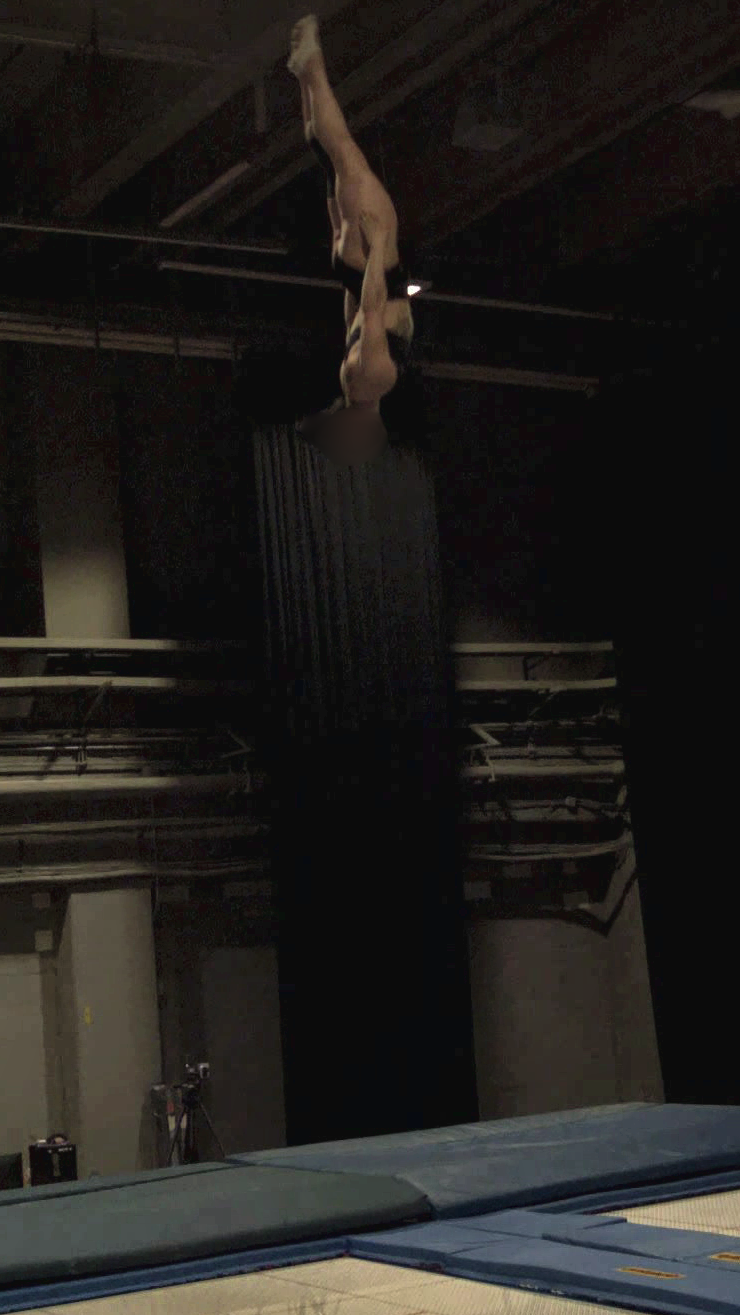}
    \end{minipage}
    \caption{Examples of images acquired from a same camera viewpoint showing the trampolinist from (a) bottom, (b) top and (c) side orientations. Images are cropped for clarity.}
\label{fig:cam_views}
\end{figure}

RePoGen \cite{purkrabek_improving_2024} is included for comparison, as their model is also a ViTPose specifically trained on synthetic data of extreme poses and uncommon viewpoints. Since trampoline athletes are always in rotation and extreme poses, and thus can be seen from top, bottom and side views from a single camera viewpoint, as shown on \cref{fig:cam_views}, we chose to train only one generic model rather than one specialized model per viewpoint, like RePoGen \cite{purkrabek_improving_2024}.

\subsection{Multi-view 3D reconstruction accuracy against baseline}

Real-world 3D reconstruction accuracy is evaluated on the MRT dataset. Once 2D poses are estimated in each view, the 3D poses are computed by triangulation, using the Pose2Sim pipeline \cite{pagnon_pose2sim_2022}. The triangulation process is an iterative optimization, trying all possible combinations of given cameras until reaching a valid triangulation. The number of used cameras must stay above the minimal number of cameras, set to three for all experiments. The reprojection error threshold to consider that a joint triangulation is valid is set to 15px, unless otherwise specified, as in the default Pose2Sim configuration file. For invalid joints, the best triangulation is kept even if the error is above the threshold to allow a fair comparison between the models.

Since the triangulation step relies primarily on camera calibration and 2D keypoints predictions, and assuming the camera calibration is accurate, this evaluation provides a meaningful robustness test for assessing the view-invariance properties of pose estimation models. In addition, it also evaluates the model's ability to accurately predict a sufficient number of joints across views to ensure complete and accurate triangulated 3D poses. This metric is more informative than COCO AP for real-world applications like sports, where the goal is to accurately retrieve a 3D pose from a minimal number of camera views.

We compare the triangulations obtained from the 2D poses estimated by the baseline ViTPose-S and by our fine-tuned version (LSPe+STP). Both predictions are compared to the ground-truth (GT) which corresponds to the triangulation of the manually annotated MRT 2D poses, using all eight views. \cref{fig:intro} provides qualitative results both in 2D and after 3D reconstruction. The example on the left shows the importance of side disambiguation for 3D reconstruction, as the skeleton ends up twisted with the baseline model, due to inconsistencies across views. The example on the right shows missing detections in 2D leading to inaccurate 3D reconstruction.

\begin{figure}[H]
    \centering
    \includegraphics[width=\columnwidth]{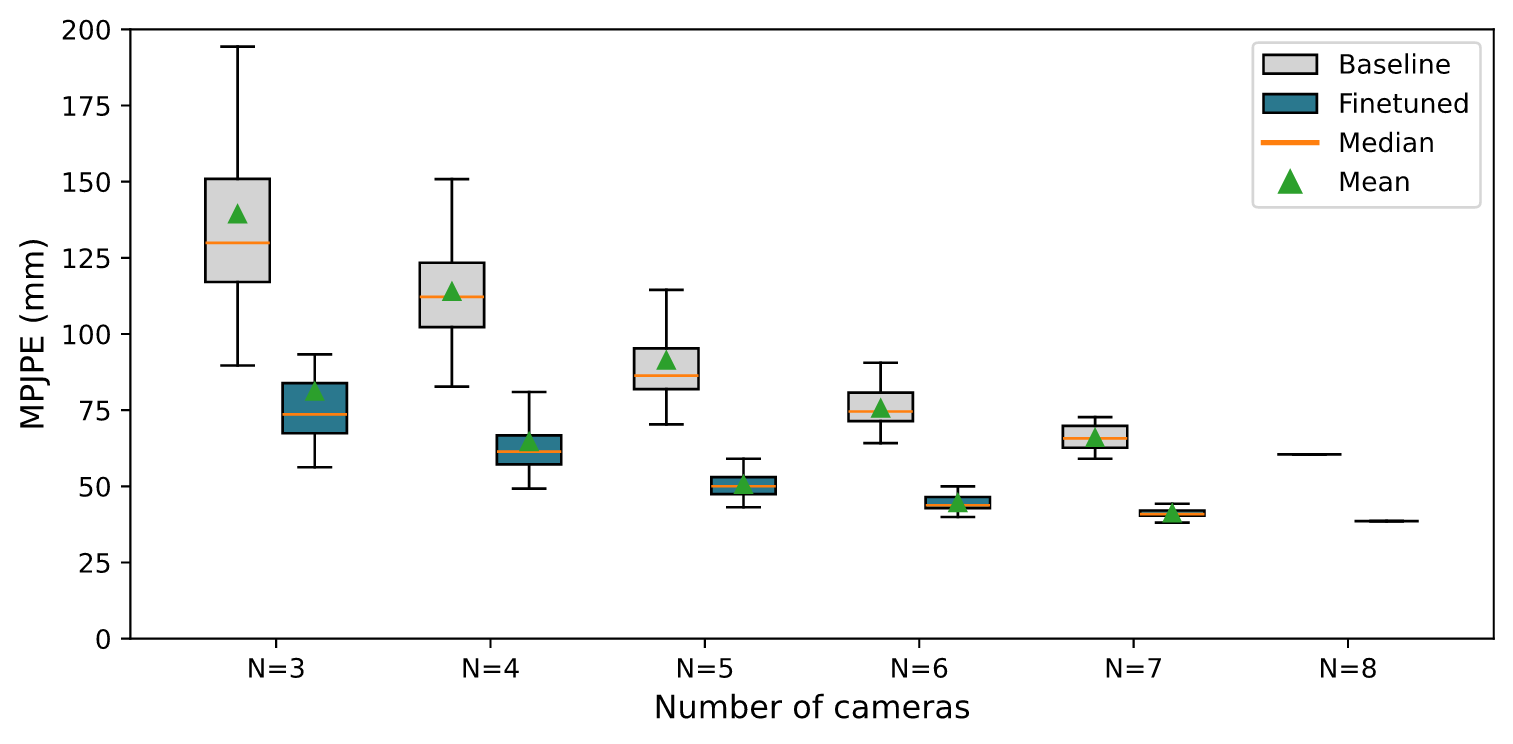}
    \caption{MPJPE distribution among camera combinations of N cameras. MPJPE is computed on 3D triangulated poses using baseline 2D predictions \textit{(gray)} or fine-tuned model 2D predictions \textit{(blue)}. Outliers are represented as circles.}
    \label{fig:3dimprovement}
\end{figure}

\cref{fig:3dimprovement} shows the distribution of the MPJPE for all combinations of N=3 to N=8 cameras. The baseline model consistently produces larger reconstruction errors. Both mean and maximal MPJPE values are smaller with our fine-tuned model. Overall, we observe an average reduction of 42.7\% (46.1 mm) in MPJPE. The lowest improvement is 21.9 mm when using all eight cameras and the highest is 58.1 mm when using only three cameras. The improvement observed when using seven or eight cameras shows that, even with multiple predictions, the fine-tuned model leads to a more accurate 3D pose reconstruction. Although one might expect that errors would be mitigated when multiple views are combined, our results suggest that triangulation tends to propagate 2D pose errors instead of mitigating them. Using only three well-selected camera views with our fine-tuned model (MPJPE=56.3 mm) is enough to reach a better accuracy than using eight camera views and the baseline model (MPJPE=61.4 mm).

\begin{figure}[h]
    \centering
    \includegraphics[width=\columnwidth]{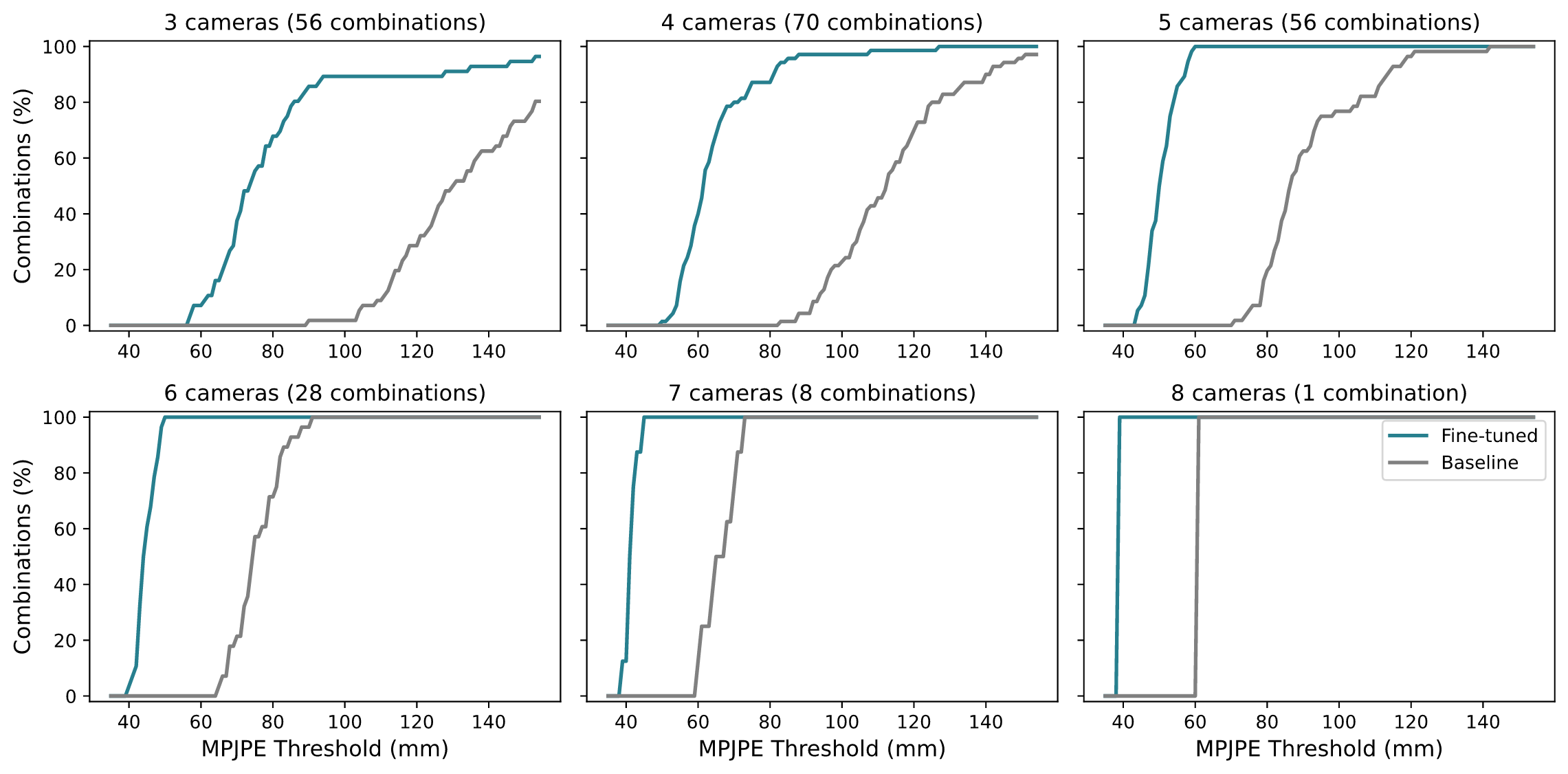}
    \caption{Percentage of camera combinations leading to triangulation error below MPJPE.  Triangulations are done using Baseline predictions \textit{(gray)} and Fine-tuned model predictions \textit{(blue)}. The percentage of joints is the average for all combinations of N cameras.} 
    \label{fig:roc_comb}
\end{figure}

\cref{fig:roc_comb} presents the percentage of camera combinations leading to triangulation below MPJPE thresholds. The MPJPE of a camera combination is the average MPJPE of all its triangulated joints. It shows that, for all numbers of cameras, the triangulations from the fine-tuned model reach a lower MPJPE with more camera combinations than triangulations from the pre-trained model. It also means that, for a given MPJPE threshold, we need less cameras using the fine-tuned model and we have more liberty in the positioning of these cameras. This is particularly interesting in sport contexts, where we might be restricted in both camera number and positioning.

\subsection{Limitations}


The ground-truth we use in both SRT and MRT comes from manual annotations. As previously mentioned, the images are challenging: even human annotators are faced with uncertainty for some joints. When the location of a joint could not be determined with sufficient confidence, it was left unannotated, and hence not considered in the metrics. Moreover, direct comparison with ViTPose's COCO AP is limited due to differences in the annotation protocol. In 3D, triangulated ground-truth joints are more reliable since the Pose2Sim triangulation process and the large number of annotated views act as a consistency check across camera views.

STP is a small set of synthetic images, limited by the size of TramPoseFit and the exclusion of a subject to avoid any form of data leakage. Our results (\cref{tab:AP}) reveal that fine-tuning exclusively on STP underperforms LSPe alone, which suggests either that STP's size is insufficient or that the domain gap between our synthetic and real data is too important. Our current evaluation cannot disentangle these two factors. All our fine-tuning experiments outperform RePoGen, showing that pose diversity and domain-specificity are essential.

The small scale of the MRT test set limits the generalization to other athletes, routines or sports. Moreover, as results are obtained with a ViTPose model, known for its heavy architecture, additional work is required to test whether these gains would transfer to lightweight real-time edge models suited for real-world deployment.

\section{Conclusion}
In this work, we address the challenge of HPE in trampoline gymnastics, an extreme benchmark where existing pipelines for motion capture and pose estimation break down, revealing gaps in current models and datasets. By introducing a realistic physics‑consistent set of acrobatic motions and releasing the full toolchain used to generate and process them, we close part of this gap and enable broader research on human pose and shape estimation in challenging, under-represented regimes. Our work shows that combining synthetic and real data helps to improve HPE in a given context, but also that the fidelity of both motion and appearance is critical when operating in highly dynamic, non-standard scenarios. Our work is a proof-of-concept, showing strong in-domain performance with our fine-tuning method. We further demonstrate that these improvements in 2D predictions translate to more reliable multi-view 3D pose reconstruction in a real-world trampoline scenario.

\section*{Acknowledgements}
This work was supported through Mitacs Accelerate program in partnership with Own the Podium. Léa Drolet-Roy gratefully acknowledges scholarship support from the Fonds de recherche du Québec – Nature et technologies (FRQNT). The authors thank the Institut national du sport du Québec, the trampoline athletes, as well as Youstina Daoud and Alexandre Naaim for their participation in data collection.

%
%
\bibliographystyle{splncs04}
\bibliography{main}
\end{document}